\documentclass{article}
\usepackage{amsmath,epsfig}

\usepackage[preprint]{spconfa4}

\newcommand{\ie}{\textit{i}.\textit{e}.}

\usepackage{booktabs}       
\usepackage{amsfonts}       
\usepackage{amsmath}
\usepackage{ dsfont }

\usepackage{paralist}
\usepackage{multirow}
\usepackage{graphicx}
\usepackage[ruled]{algorithm2e}
\SetKwComment{Comment}{$\triangleright$\ }{}

\copyrightnotice{978-1-6654-3864-3/21/\$31.00 ©2021 IEEE}

\let\OLDthebibliography\thebibliography
\renewcommand\thebibliography[1]{
  \OLDthebibliography{#1}
  \setlength{\parskip}{0pt}
  \setlength{\itemsep}{0pt plus 0.3ex}
}

\begin{document}\sloppy

\def\x{{\mathbf x}}
\def\L{{\cal L}}

\title{Universal Adversarial Training with Class-Wise Perturbations}
%
\name{Philipp Benz$^*$\thanks{$^*$ Equal contribution}, Chaoning Zhang$^*$, Adil Karjauv, In So Kweon}
\address{Korea Advanced Institute of Science and Technology (KAIST)\\
\small{\texttt{\{pbenz, iskweon77\}@kaist.ac.kr}, \texttt{\{chaoningzhang1990, mikolez\}@gmail.com}}}

\maketitle

\begin{abstract}
Despite their overwhelming success on a wide range of applications, convolutional neural networks (CNNs) are widely recognized to be vulnerable to adversarial examples. This intriguing phenomenon led to a competition between adversarial attacks and defense techniques. So far, adversarial training is the most widely used method for defending against adversarial attacks. It has also been extended to defend against universal adversarial perturbations (UAPs). The SOTA universal adversarial training (UAT) method optimizes a single perturbation for all training samples in the mini-batch. In this work, we find that a UAP does not attack all classes equally. Inspired by this observation, we identify it as the source of the model having unbalanced robustness. To this end, we improve the SOTA UAT by proposing to utilize class-wise UAPs during adversarial training. On multiple benchmark datasets, our class-wise UAT leads superior performance for both clean accuracy and adversarial robustness against universal attack. 
\end{abstract}
\section{Introduction}
\label{sec:intro}

Convolutional neural networks (CNNs) have achieved significant progress on vision applications~\cite{zhang2020deepptz,zhang2020resnet,zhang2019revisiting}. However, CNNs are widely known to be vulnerable to adversarial examples, \ie\ images perturbed by imperceptible malicious perturbations. This intriguing phenomenon has led to active research on model robustness~\cite{akhtar2018threat,benz2020data,benz2020batch,benz2020revisiting,benz2020robustness,feng2020adversarial}. To mitigate the adversarial effects of such attacks, numerous works have investigated different defense techniques. Most defense techniques, however, have been proven to fail by later works~\cite{athalye2018obfuscated}. To our best knowledge, adversarial training is so far the only method that has not been broken by strong white-box attacks, such as PGD~\cite{madry2017towards}. With the FGSM adversarial training~\cite{goodfellow2014explaining} and PGD adversarial training~\cite{madry2017towards} as the baseline, the progress of adversarial training can be summarized into two aspects: (a) addressing the balance between accuracy and robustness~\cite{benz2020robustness,zhang2019theoretically} and (b) increasing the training speed~\cite{shafahi2019adversarial,wong2020fast}.

Most prior works focused on adversarial training against image-dependent adversarial attacks instead of universal attacks.~\cite{moosavi2017universal} first demonstrates that there exists a single perturbation that can fool the target network for most images. Due to its image-agnostic property, it is often called universal adversarial perturbation (UAP). The UAP can be generated beforehand and applied to attack the target model directly~\cite{moosavi2017universal,zhang2020understanding}, which makes it more suitable for real-world attacks. Thus, improving model robustness against universal attack is of high practical relevance. Currently, there are multiple works~\cite{mummadi2019defending,shafahi2020universal} that have attempted to extend adversarial training to defend against UAPs. Our work is most similar to universal adversarial training (UAT) proposed in~\cite{shafahi2020universal}. It updates the neural network parameters and the universal perturbation concurrently, which is empirically found to be fast yet effective~\cite{shafahi2020universal}. It adopts a single UAP during the training to make the neural network robust against UAPs. We find that the generated UAP often does not attack images equally for each class. Given that UAPs contain dominant semantic features of a certain class~\cite{zhang2020understanding}, this class-wise imbalance~\cite{benz2020robustness} is somewhat expected. Moreover, since only a single UAP is adopted for all samples during adversarial training, there is inevitably a lack of diverse perturbation directions. For mitigating the above concerns, we propose class-wise UAT, \ie\ replacing the single UAP with class-wise UAPs during training. Our class-wise UAT outperforms the SOTA UAT~\cite{shafahi2020universal} by a considerable margin for both clear accuracy and adversarial robustness against universal attack.

\section{Related Work}
\textbf{Universal Adversarial Perturbations.}
After the initial discovery of UAPs in~\cite{moosavi2017universal}, multiple methods for generating them have been developed. The classical UAP algorithm in~\cite{moosavi2017universal} accumulates image-dependent perturbations through applying the image-dependant DeepFool attack~\cite{moosavi2016deepfool} iteratively. Adopting a generative network for crafting UAPs has also been investigated in~\cite{poursaeed2018generative,hayes2018learning}. A UAP attacking images from all classes without intended class discrimination can easily raise suspicion. To make the generated UAP more stealthy, class-discriminative UAPs have been investigated in~\cite{zhang2019cd-uap} to attack images from a predefined group of classes while minimizing the adversarial effect on other classes.~\cite{benz2020double} further proposed a double targeted UAP that targets the class on both the source and sink side as well as exploring its physical attack capabilities with an adversarial patch. For explaining the existence of UAPs,~\cite{moosavi2017analysis} shows that its existence can be attributed to the positive curvature of the decision boundary. It has been shown in~\cite{jetley2018friends} that the input directions, that the classifier is vulnerable to, aligns well with the input directions that are useful for classification.~\cite{Mopuri2017datafree,mopuri2018generalizable,reddy2018ask} investigate data-free UAPs and~\cite{zhang2020understanding} achieves the first data-free targeted UAP.~\cite{zhang2020understanding} explains the UAP from the mutual influence of images and perturbations, revealing that UAPs have the semantic features of a certain class and images behave like noise to the UAP. This intriguing phenomenon can be partially explained by the finding in~\cite{zhang2021universal} that deep classifier is sensitive to high frequency content. studies the UAP through the lens of deep steganography. We refer the readers to~\cite{zhang2021survey} for a comprehensive survey on universal adversarial attack.

\textbf{Defense Against Universal Attack.}
Adversarial training is the most widely used defense against adversarial attacks~\cite{goodfellow2014explaining,madry2017towards}. Most other defense or detection methods have been shown to fail and adversarial training remains, to our best knowledge, the only defense method that has not been broken by white-box attacks~\cite{athalye2018obfuscated}. Numerous works have investigated adversarial training for defending against image-dependent adversarial attacks~\cite{goodfellow2014explaining,madry2017towards,shafahi2019adversarial,zhang2019you,wong2020fast}. Adversarial training has also been extended to defend against UAPs by fine-tuning the model parameters on images perturbed by UAPs generated through pre-computation~\cite{moosavi2017universal}, a generative model~\cite{hayes2018learning}, which unfortunately only slightly increases the robustness against UAP. Note that this is somewhat expected because the applied UAP is fixed, which is different from normal adversarial training that dynamically changes the perturbation during training~\cite{goodfellow2014explaining,madry2017towards}. The challenge of adopting the same procedure as normal adversarial training for UAP is that crafting UAPs~\cite{moosavi2017universal} is more time-consuming. To address this problem,~\cite{mummadi2019defending} has proposed shared adversarial training to generate a UAP on the fly. However, it is still 20 times slower than the normal training because it generates UAPs similar to PGD adversarial training~\cite{madry2017towards}. To our knowledge, universal adversarial training (UAT)~\cite{shafahi2020universal} is currently the state-of-the-art approach that solves this problem elegantly by concurrently updating the networks and perturbation, which makes it add no extra computation overhead compared with normal training. The efficacy and efficiency of this approach have also been shown in fast adversarial training~\cite{wong2020fast}. Thus, in this work, we adopt UAT~\cite{shafahi2020universal} as the baseline and improve it by adopting class-wise UAPs to increase the diversity of the UAP during training. 

\section{Background and Motivation}
Universal Adversarial Perturbations (UAPs)~\cite{moosavi2017universal} fool most images with one single perturbation. Assuming a classifier $\hat{C}_\theta$ parameterized through weights $\theta$ (from here on omitted) as well as a dataset $\mathcal{X}$ from which samples $x$ can be drawn, the objective to craft a single UAP perturbation $\delta$ is:
\begin{align*}
    &\hat{C}(x+\delta) \neq \hat{C}(x) \text{ for most } x \sim \mathcal{X} \\
    &\text{subject to } ||\delta|| \leq \epsilon.
\end{align*}
Here $\epsilon$ indicates an upper bound for the permissible perturbation magnitude. The perturbation magnitude is calculated via the $l_p$-norm indicated by $||\cdot||$. The classical UAP~\cite{moosavi2017universal} adopted the image-dependent attack method DeepFool~\cite{moosavi2017universal} to iteratively accumulate the final perturbation.~\cite{shafahi2020universal,zhang2020understanding} argued, that this process is slow and proposed faster and more efficient UAP algorithms. We adopt the state-of-the-art algorithm proposed in~\cite{zhang2020understanding} to craft UAPs, which is shown in Algorithm~\ref{alg:uap_crafting}.

\begin{algorithm}[!htbp]
    \SetAlgoLined
    \DontPrintSemicolon
    \SetKwInput{KwInput}{Input}
    \SetKwInput{KwOutput}{Output}
    \SetKwFunction{FOptim}{Optim}
    \SetKwFunction{FClamp}{Clamp}
    \KwInput{Training Data $\mathcal{X}$, Classifier $\hat{C}$, Loss function $\mathcal{L}$, Optimizer \FOptim, Number of iterations $I$, perturbation magnitude $\epsilon$}
    \KwOutput{Perturbation vector $\delta$}
    $\delta \leftarrow 0$ \Comment*[r]{Initialization}
    \For {iteration $=1, \dots, I$}{
        $B \sim \mathcal{X}$ \Comment*[r]{Random batch sampling}
        $g_\delta \leftarrow \underset{x \sim B}{\mathds{E}} [\nabla_{\delta} \mathcal{L}$] \Comment*[r]{Calculate gradient} 
        $\delta \leftarrow$ \FOptim{$g_\delta$} \Comment*[r]{Update} 
        $\delta \leftarrow $ \FClamp{$\delta, -\epsilon, \epsilon$} \Comment*[r]{Norm projection}
        }
\caption{UAP Crafting Algorithm}
\label{alg:uap_crafting}
\end{algorithm}

\begin{figure}[t]
    \centering
    \includegraphics[width=\linewidth]{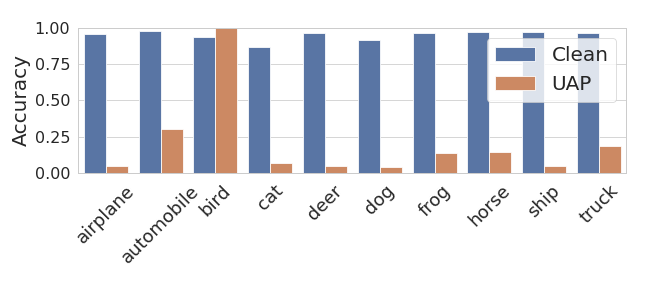}
    \includegraphics[width=\linewidth]{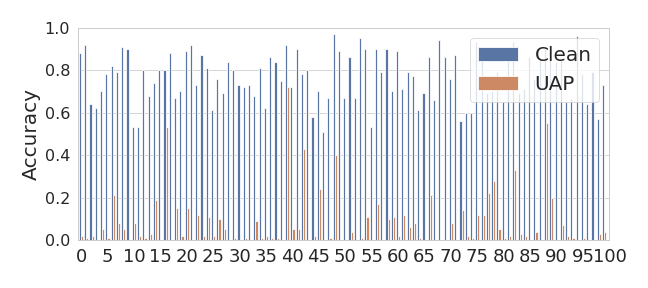}
    \caption{Class-wise accuracies without (``Clean") and with UAP (``UAP") applied. The plots are shown for a standard ResNet-18 on CIFAR-10 (top) and CIFAR-100 (bottom).}
    \label{fig:cifar_uap}
\end{figure}

Inspired by adversarial training for image-dependent adversarial examples~\cite{goodfellow2014explaining}, universal adversarial training (UAT)~\cite{shafahi2020universal} had been proposed as a defense against UAPs. 
In the following, we highlight that UAPs do not attack images from all classes equally, which finally can lead to some inefficiencies in UAT. For this, we plot the class-wise accuracies of a ResNet-18 trained on CIFAR for all samples perturbed through a UAP. The UAP was crafted with Algorithm~\ref{alg:uap_crafting} for $\epsilon=8$ and $2000$ iterations with the ADAM optimizer. The plots in Figure~\ref{fig:cifar_uap} show that there exists a discrepancy in the class-wise accuracies. For example for the model trained on the CIFAR-10 training dataset and evaluated on the CIFAR-10 validation dataset, the class-wise accuracies overall decrease after the UAP is applied, except for the class ``bird". Note that before applying the UAP the class-wise discrepancy also exists but much less significant. The results show that the UAP does not attack all classes equally. The same observation can be made for the ResNet-18 trained and evaluated on CIFAR-100. Among the 100 classes, numerous classes remain with a high accuracy indicated by the orange spikes for certain classes. 

\begin{figure}[!htbp]
    \centering
    \includegraphics[width=\linewidth]{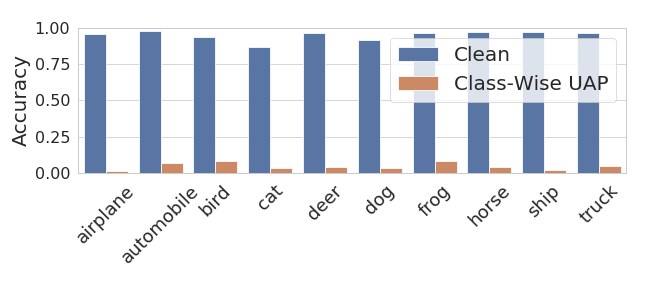}
    \includegraphics[width=\linewidth]{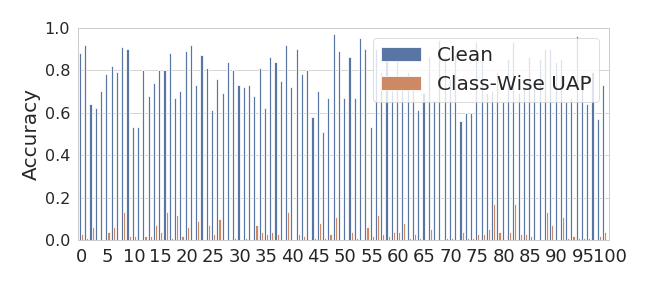}
    \caption{Class-wise accuracies without (``Clean") and with class-wise UAP (``Class-Wise UAP") applied to the images. The plots are shown for a standard ResNet-18 on CIFAR-10 (top) and CIFAR-100 (bottom).}
    \label{fig:cifar_class_wise_uap}
\end{figure}

By contrast, crafting one UAP for the samples of each class separately, hence resulting in $N$ number of UAPs, where $N$ is the number of classes, the class-wise accuracies of all classes can be significantly decreased. We term such perturbations class-wise UAPs. Figure~\ref{fig:cifar_class_wise_uap} depicts the increased severity of the class-wise UAP. The class-wise accuracies for the ResNet-18 model trained and evaluated on CIFAR-10 are all below or around $10\%$ point. The same trend is observed for the ResNet-18 model trained on CIFAR-100. This improved effectiveness in terms of universal adversarial attack motivates us to use class-wise UAPs instead of a single UAP during universal adversarial training. Class-wise UAPs can force the learned UAPs to be more diverse, thus likely to lead to a more effective UAT.

\section{Universal Adversarial Training with Class-Wise UAPs}
We define a class-wise UAP $\nu$ as a single perturbation which has the objective to fool most of the samples $x_c \sim \mathcal{X}_c$ with ground truth class $c$ while obeying the magnitude constraint $\epsilon$. $\mathcal{X}_c$ indicates the distribution of all samples with ground truth class $c$. More formally: 
\begin{align*}
    &\hat{C}(x_c+\nu) \neq \hat{C}(x_c) \text{ for most } x_c \sim \mathcal{X}_c \\
    &\text{subject to } ||\nu|| \leq \epsilon.
\end{align*}
Note that in practice class-wise UAPs cannot be used as an attack, since during the inference stage the ground truth class is unknown, hence the corresponding class-wise UAP cannot be assigned to its respective sample. It is also worth mentioning that unlike class-discriminate UAP~\cite{zhang2019cd-uap}, class-wise UAP still attacks other classes very effectively. Nevertheless, the class-wise UAPs are more diverse thus constitutes a better solution to harden a model against universal attack. 

In Algorithm~\ref{alg:uat} we adapt the UAT algorithm introduced in~\cite{shafahi2020universal} to leverage class-wise UAPs. The resulting new algorithm is termed class-wise UAT. 

\begin{algorithm}[t]
    \SetAlgoLined
    \DontPrintSemicolon
    \SetKwInput{KwInput}{Input}
    \SetKwInput{KwOutput}{Output}
    \SetKwFunction{FOptimnet}{Optim$_\theta$}
    \SetKwFunction{FOptimuap}{Optim$_\nu$}
    \SetKwFunction{FClamp}{Clamp}
    \KwInput{Training data $\mathcal{X}$, Network weights $\theta$, Loss function $\mathcal{L}$, Network optimizer \FOptimnet, UAP optimizer \FOptimuap, Perturbation magnitude $\epsilon$}
    \KwOutput{Robust Classifier $\hat{C}_\theta$}
    $\nu \leftarrow 0$ \Comment*[r]{Class-wise UAPs}
    \For {epoch $=1, \dots E$}{
        \For {batch $B$ in $\mathcal{X}$}{
            \tcc{Apply class-wise UAP}
            \For {$c$ in number of classes $N$}{
                $x_c \leftarrow$ Samples with ground truth $c$ \\
                $\nu_c \leftarrow$ $c$-th class-wise UAP \\
                $x'_c \leftarrow  x_c + \nu_c$ 
            }
            \tcc{Simultaneously calculate gradients}
            $g_\nu, g_\theta \leftarrow \underset{x \sim B}{\mathds{E}} [\nabla_{\nu, \theta} \mathcal{L}(\hat{C}(x'),y_{gt})$] \\
            $\theta \leftarrow$ \FOptimnet{$g_\theta$} \Comment*[r]{Update Network} 
            $\nu \leftarrow$ \FOptimuap($-g_\nu$) \Comment*[r]{Update UAPs} 
            $\nu \leftarrow $ \FClamp{$\nu, -\epsilon, \epsilon$} \Comment*[r]{Norm projection}
            }
        }
\caption{Universal Adversarial Training with Class-Wise UAPs}
\label{alg:uat}
\end{algorithm}

\section{Results}
Following, we will present the results for improved universal adversarial training with class-wise UAP. For all our experiments, we use a ResNet-18~\cite{he2016deep} or a 10 times wider variant Wide-ResNet (WRN)~\cite{zagoruyko2016wide}. Following~\cite{shafahi2020universal}, we consider $l_\infty$ perturbations and set $\epsilon$ to $8/255$ for images in the range $[0,1]$.

\subsection{Comparison with SOTA UAT}
We compare our method with the results reported in~\cite{shafahi2020universal} for the Wide-ResNet. The results on CIFAR10 are available in Table~\ref{tab:comparison_uat}. Our approach outperforms the previous SOTA method UAT~\cite{shafahi2020universal} by a significant margin. Improved UAT with class-wise UAPs can improve the robustness of the model to UAPs by $3\%$ point. It is also noticeable that not only the universal adversarial robustness but also the clean accuracy is improved. Moreover, for our proposed class-wise UAT method, the accuracy drop under the UAP attack is only $0.1\%$ point, which suggests that our proposed UAT training method results in a model that is almost immune to the UAP attack.

\begin{table}[t]
\centering
\caption{Comparison with UAT~\cite{shafahi2020universal} on CIFAR-10.}
\label{tab:comparison_uat}
\begin{tabular}{cccccc}
\toprule
Method                                 & Clean &  UAP   \\ \midrule
UAT (ADAM)~\cite{shafahi2020universal} & 94.3  & 91.6   \\
UAT (FGSM)~\cite{shafahi2020universal} & 93.5  & 91.8   \\ 
Class-wise UAT (Ours)                  & \textbf{94.9} & \textbf{94.8}   \\
\bottomrule
\end{tabular}
\end{table}

\subsection{CIFAR Results}
In the following, we present a more extended comparison of our improved UAT to the adversarial training techniques using PGD and UAT. Since the authors of UAT~\cite{shafahi2020universal} did not open-source their code, we implemented UAT by ourselves. For adversarial training with PGD, we use $7$ update steps, update the perturbations with the $l_\infty$ norm, allow a maximum perturbation magnitude of $\epsilon=8/255$ and calculate the step size as $2.5 \epsilon / \text{steps}$. The results are available in Table~\ref{tab:results_cifar10}. Among the adversarial training variants, PGD-adversarial training shows the least robustness against UAPs, which is somewhat expected, since it is optimized on image-dependent adversarial perturbations. PGD-adversarial training also exhibits the lowest clean accuracy among the tested adversarial training techniques. Compared to UAT~\cite{shafahi2020universal}, our class-wise UAT outperforms by a significant margin for both models. Specifically, for ResNet-18 training with class-wise UAPs can improve the performance by $4.5\%$ point, while for the Wide-ResNet, the performance improves by $5\%$ point. 

\begin{table}[t]
\centering
\caption{Comparison of different adversarial training techniques on CIFAR-10 for ResNet18 (top) and Wide-ResNet (bottom).}
\label{tab:results_cifar10}
\begin{tabular}{cccccc}
\toprule
Model                & Clean & UAP  \\ \midrule
ResNet-18            & 94.9  & 20.9 \\
ResNet-18 (PGD)      & 83.8  & 82.6 \\
ResNet-18 (UAT)      & 93.7  & 88.5 \\
ResNet-18 (Class-wise UAP)   & 93.6  & 93.0 \\ \midrule
WRN-28-10                  & 96.2  & 20.2 \\
WRN-28-10 (PGD)            & 86.8  & 86.1 \\
WRN-28-10 (UAT)            & 94.4  & 89.8 \\
WRN-28-10 (Class-wise UAT) & 94.9  & 94.8 \\
\bottomrule
\end{tabular}
\end{table}

We further evaluate the different adversarial training variants on CIFAR-100. The results in Table~\ref{tab:results_cifar100} resemble those in Table~\ref{tab:results_cifar10}. Overall, our proposed approach outperforms UAT~\cite{shafahi2020universal} in terms of clean accuracy and robust accuracy under universal attack. 

\begin{table}[t]
\centering
\caption{Comparison of different adversarial training techniques on CIFAR-100 for ResNet18 (top) and Wide-ResNet (bottom).}
\label{tab:results_cifar100}
\begin{tabular}{cccccc}
\toprule
Model                      & Clean & UAP \\ \midrule
ResNet-18                  & 76.5 & 6.6  \\
ResNet-18 (PGD)            & 56.2 & 54.7 \\
ResNet-18 (UAT)            & 69.7 & 66.8 \\ 
ResNet-18 (Class-wise UAT) & 70.3 & 68.2 \\ \midrule
WRN-28-10                  & 81.9 & 5.4  \\
WRN-28-10 (PGD)            & 60.5 & 59.8 \\
WRN-28-10 (UAT)            & 76.0 & 71.8 \\
WRN-28-10 (Class-wise UAT) & 77.1 & 74.0 \\
\bottomrule
\end{tabular}
\end{table}

\subsection{ImageNet-10 Results}
Previously, we have shown the efficacy of our proposed approach on CIFAR. Here, we evaluate its efficacy on higher resolution images. We randomly select a subset of 10 classes from the ImageNet dataset (``drake", ``beaver", ``ballpoint", ``beach wagon", ``palace", ``red wolf", ``water buffalo", ``barbershop", ``chain-link fence", ``banana"). The result resembles that on the CIFAR dataset. For example, on WRN-50, our class-wise UAT significantly improves the robustness against UAP from $81.3\%$ to $90.4\%$.

\begin{table}[t]
\centering
\caption{Comparison of different adversarial training techniques on ImagNet-10 for ResNet-18 (top) and Wide-ResNet (bottom)}
\label{tab:results_imagenet10}
\begin{tabular}{cccccc}
\toprule
Model                      & Clean & UAP  \\ \midrule
ResNet-18                  & 92.8 & 49.0  \\
ResNet-18 (UAT)            & 89.8 & 80.2  \\
ResNet-18 (Class-wise UAT) & 89.2 & 85.0  \\ \midrule
WRN-50                     & 97.6 & 49.6  \\
WRN-50 (UAT)               & 89.0 & 81.3 \\
WRN-50 (Class-wise UAT)    & 91.0 & 90.4 \\
\bottomrule
\end{tabular}
\end{table}

\subsection{Analysis and visualization}
\begin{figure}[t]
    \centering
    \includegraphics[width=\linewidth]{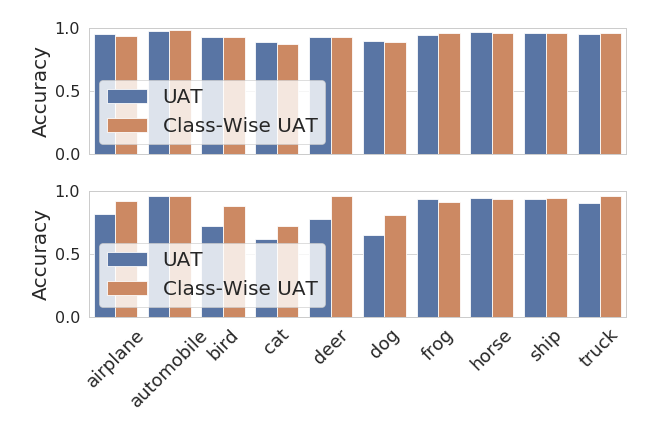}
    \caption{Class-wise accuracies without (top) and with a UAP applied (bottom) for a ResNet-18 model trained with UAT and our proposed class-wise UAT on CIFAR-10.}
    \label{fig:class_wise_acc_robust}
    
\end{figure}

We investigate the class-wise behavior of a robust model trained with our proposed class-wise UAT. In Figure~\ref{fig:class_wise_acc_robust}, we show the class-wise accuracies without and with the application of a UAP for a ResNet-18 trained with UAT~\cite{shafahi2020universal} and our proposed class-wise UAT.
It can be observed that the model trained with UAT~\cite{shafahi2020universal} shows more unbalanced class-wise performance, with the accuracies of some vulnerable classes, such as ``bird", ``cat", ``deer" and ``dog", being relatively low, while the others are relatively high. A similar trend can be observed for our proposed class-wise UAT, however, the robustness of those vulnerable classes increases by a large margin. Thus, our class-wise UAT not only contributes to a boosted overall robustness but also realizes a more balanced class-wise robustness against universal attack. 

\begin{figure}[!htbp]
    \centering
    \includegraphics[width=\linewidth]{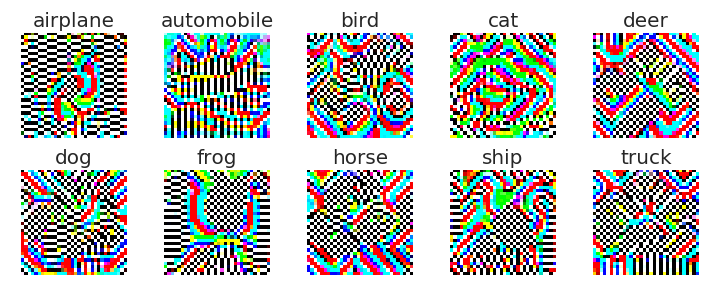}
    \includegraphics[width=\linewidth]{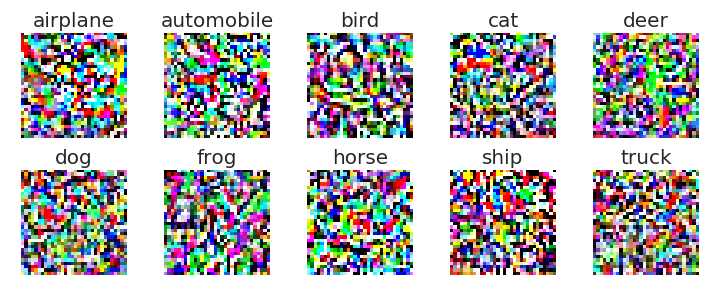}
    \caption{Class-wise UAPs generated for samples from the different ground truth classes. The class-wise UAPs were generated for a standard ResNet-18 (top) and an adversarially trained ResNet-18 with the proposed UAT with class-wise UAPs (bottom).}
    \label{fig:vis_pert}
\end{figure}

Figure~\ref{fig:vis_pert} visualizes the amplified versions of our class-wise UAPs for a standard ResNet-18 and a robustified ResNet-18 obtained with our proposed method. The class-wise UAPs for a standard model appear to be more structured with repeating patterns. The patterns for the robust model appear to be more locally smooth, which echos with the finding in~\cite{ilyas2019adversarial} that adversarially trained models are more dependent on structured features. 

\section{Conclusion}
In this work, we show that the UAP does not attack images from all classes equally and we identify the SOTA UAT algorithm that only adopts a single perturbation during training as a source for unbalanced robustness. To this end, we propose class-wise UAT that outperforms the existing SOTA UAT by a large margin for both clean accuracy and adversarial robustness against universal attack. Due to the enhanced robustness of vulnerable classes, our class-wise UAT also contributes to a more balanced  class-wise robustness.

\bibliographystyle{IEEEbib}
\bibliography{bib_mixed}

\end{document}